\documentclass[acmsmall]{acmart}

\renewcommand\footnotetextcopyrightpermission[1]{}
\def\BibTeX{{\rm B\kern-.05em{\sc i\kern-.025em b}\kern-.08emT\kern-.1667em\lower.7ex\hbox{E}\kern-.125emX}}

\setcopyright{none}

\makeatletter
\def\runningfoot{\def\@runningfoot{}}
\def\firstfoot{\def\@firstfoot{}}
\makeatother 

\begin{document}

\title{Gaze-enhanced Crossmodal Embeddings for Emotion Recognition}

\author{Ahmed Abdou}
\affiliation{%
  \institution{Technical University of Munich}
  \country{Germany}
}
\email{ahmed.abdou@tum.de}

\author{Ekta Sood}
\affiliation{%
  \institution{University of Stuttgart}
  \country{Germany}
}
\email{ekta.sood@vis.uni-stuttgart.de}

\author{Philipp Müller}
\affiliation{%
  \institution{German Research Center for Artificial Intelligence}
  \country{Germany}
}
\email{philipp.mueller@dfki.de}

\author{Andreas Bulling}
\affiliation{%
  \institution{University of Stuttgart}
  \country{Germany}
}
\email{andreas.bulling@vis.uni-stuttgart.de}

\renewcommand{\shortauthors}{Ahmed Abdou et al.}

\begin{abstract}
Emotional expressions are inherently multimodal -- integrating facial behavior, speech, and gaze -- but their automatic recognition is often limited to a single modality, e.g. speech during a phone call. 
While previous work proposed crossmodal emotion embeddings to improve monomodal recognition performance, despite its importance, an explicit representation of gaze was not included.
We propose a new approach to emotion recognition that incorporates an explicit representation of gaze in a crossmodal emotion embedding framework. 
We show that our method outperforms the previous state of the art for both audio-only and video-only emotion classification on the popular One-Minute Gradual Emotion Recognition dataset.
Furthermore, we report extensive ablation experiments and provide detailed insights into the performance of different state-of-the-art gaze representations and integration strategies. 
Our results not only underline the importance of gaze for emotion recognition but also demonstrate a practical and highly effective approach to leveraging gaze information for this task.
\end{abstract}





\maketitle
\section{Introduction}
Automatic recognition of emotional expressions is an inherently multimodal task~\cite{pantic2005affective,zeng2008survey}. 
Many state-of-the-art approaches combine information extracted from multiple modalities, e.g. expressions in a persons' face with speech-based features~\cite{wu2014survey,rouast2019deep,schoneveld2021leveraging}.
While these approaches consistently outperform uni-modal alternatives~\cite{rouast2019deep,d2015review} they share one crucial limitation: They require all modalities to be simultaneously present at both training and test time.
However, this assumption rarely holds in application scenarios for emotion recognition.
For example, emotion recognition systems have to rely on audio only at test time if used for telephone-based screening in the medical domain~\cite{konig2021measuring}, or when users are outside the field of view of the camera in a video conference.
In contrast, if a user is silent or strong background noise overlays a conversation, video analysis might be the only usable modality.
To combine the advantages of multimodal training with the flexibility of only requiring a subset of these modalities at test time, recent work proposed \textit{cross-modal emotion embeddings}~\cite{han2019emobed}.
In this approach, a helper modality (e.g. video) is used during training time to improve the latent representation of a second modality (e.g. audio).
This approach was shown to improve performance when only using the second modality (audio) at test time.

At the same time, a large body of work has demonstrated the close link between gaze and emotional expressions~\cite{emery2000eyes,adams2003perceived,adams2005effects,keltner1995signs,liang2021emotional}.
For example, gaze aversion was shown to impair the perception of anger and happiness~\cite{adams2005effects,bindemann2008eye}, and  embarrassment is connected to more downward gaze than amusement~\cite{keltner1995signs}.
Despite the importance of gaze, relatively few works have studied emotion recognition based on gaze location and pupil size ~\cite{aracena2015neural} or combined gaze with other channels of affective information~\cite{o2019eye,van2019emotion,alhargan2017multimodal}.
While the performance improvements demonstrated by these approaches underline the importance of integrating gaze into emotion recognition systems, they either rely on video information only~\cite{van2019emotion} or assume both video-based gaze features and speech input to be available both at training and test time~\cite{o2019eye,alhargan2017multimodal}.
In particular, so far gaze has not been integrated in multi-modal (i.e. video and audio) emotion recognition approaches that only require unimodal (i.e. either video or audio) data at test time.

In this work, we are first to propose to include an explicit representation of gaze in a crossmodal learning framework for emotion recognition.
More specifically, our approach builds on the crossmodal emotion embeddings model (EmoBed) by~\citet{han2019emobed}.
We augment the visual pipeline of EmoBed with a state-of-the-art gaze feature representation for gaze-based emotion recognition~\cite{o2019eye}.
The resulting novel feature representation of the video modality acts as a helper modality during training for speech-only testing, or is supported by speech during training when performing video-only test evaluations.
With a model-level fusion strategy of gaze and facial features, we achieve a F1 score of 45.0 for video-only testing on the One-Minute Gradual Emotion Recognition dataset~\cite{barros2018omg}, clearly improving over the previous state of the art by~\cite{han2019emobed}.
For audio-only testing, we reach 43.4 F1 with an early fusion approach for gaze integration at training time, also outperforming the previous state of the art~\cite{han2019emobed}.  \footnote{Code and other supporting material is made publicly available at \url{https://perceptualui.org/publications/abdou22_etra/}}

The specific contributions of our work are threefold:
First, we present a novel state-of-the-art crossmodal learning approach for emotion recognition that, for the first time, makes use of an explicit representation of gaze.
Second, we conduct experiments on the One-Minute Gradual Emotion Recognition dataset~\cite{barros2018omg}, improving over the previous state of the art both for video-only as well as audio-only testing.
Third, we perform extensive ablation experiments and report results for different gaze integration strategies, including early versus model-level fusion, as well as different state-of-the-art gaze feature representations~\cite{o2019eye,van2019emotion}.

\section{Related Work}

Our work is related to 1) the connection between gaze and displays of emotions, 2) gaze-based automatic emotion recognition, as well as 3) multimodal approaches to emotion recognition.
\subsection{Gaze and Emotions}

Eye gaze behavior (coupled with other modalities such as facial expressions) has been studied widely in psychology of emotion expression and perception~\cite{itier2009neural,milders2011detection,macrae2002you,wells2016identification, kimble1980gaze}. 
Gaze can reveal information about the users attention and intentions~\cite{itier2009neural}, and specific eye movement behaviors (e.g., gaze direction) coupled with facial movements~\cite{adams2003perceived, bindemann2008eye, wieser2012faces} are relevant when expressing and perceiving specific kinds of emotion classes. 
With such facial/eye region expressions, the additional use of perceived gaze direction (either directed or averted) has been shown to better assist humans to distinguish between emotion classes~\cite{emery2000eyes,adams2003perceived,milders2011detection,macrae2002you, liang2021emotional}.

Gaze direction has been shown to be associated with "the underlying behavioral intent (approach-avoidance) communicated by an emotional expression''~\cite{adams2005effects}.
~\citet{milders2011detection} showed that the gaze direction of another person can affect your emotion recognition accuracy and the intensity by which you perceive the emotional stimuli. Moreover, results showed that averted gaze is a useful feature when detecting fear over happiness or anger and the exact opposite with a direct gaze.
In addition,~\cite{keltner1995signs} indicated that when humans experience embarrassment, they first avert their gaze and then subsequently additional expressions occur, such as shifting eye, abnormal speech sounds, and smiling.

\subsection{Gaze-based Emotion Recognition}
Given the strong link between gaze and emotions, a large body of research has explored the use of gaze for automatic emotion recognition.
For example,~\citet{jaques2014predicting} explored different machine learning algorithms
for gaze-based recognition of two emotional states, curious and bored, in an e-learning environment. They found that while predicting curiosity was not possible above chance level, predicting boredom showed more promise. Similarly,~\cite{aracena2015neural} trained a shallow feed-forward neural network to predict positive, negative, or neutral emotion classes. To improve performance, they suggested to include additional modalities and gaze features in future work. \citet{o2019eye} explored the use of a larger gaze feature set to train an LSTM network for the task of continuous affect prediction from the RECOLA dataset~\cite{ringeval2013introducing}. 
They found that their model performed better for arousal prediction when trained on gaze features. 
A number of works have provided strong evidence for the benefit of using gaze for emotion recognition. 
~\citet{anwar2019real} proposed a method
to predict seven basic emotion classes from facial action units and gaze with 93\% accuracy.
Similarly, \citet{alhargan2017multimodal} explored the use of speech and gaze features for affect recognition in a gaming environment and reported top performance when combining both modalities.
Their results further showed that gaze features were even more helpful than speech.
In a study by \citet{o2018affective}, the addition of eye gaze features to speech yielded an improvement of 19.5\% for valence prediction and 3.5\% for arousal prediction.
An opposing pattern was found by
~\citet{o2019eye} who employed pupillometry and gaze features for valence and arousal estimation on the RECOLA dataset 
\cite{ringeval2013introducing}. 
While eye-based features did not perform well for arousal prediction,
the best performance was achieved when combining eye-based features and speech features.
~\citet{van2019emotion} presented a neural network approach for emotion recognition based on facial expression and gaze and
showed clear improvements when using gaze features to representations of facial expressions.
Despite all of these works, to the best of our knowledge, no emotion recognition approach incorporating gaze with both facial expressions and speech has been proposed.

\subsection{Multimodal Emotion Recognition}
Due to their ubiquity, most works on multimodal emotion recognition have focused on combining audio and video~\cite{wu2014survey,rouast2019deep}, but how to combine them remains an open question.
In early fusion, inputs or raw feature representations are merged before they are fed into a joint network~\cite{rozgic2012emotion,chen2017multimodal}.
In model-level fusion, each modality is processed by a dedicated network before both intermediate feature representations are merged and then passed through a joint network~\cite{chen2014emotion,ringeval2015prediction}.
Finally, in late fusion, modalities are fused on the level of predictions~\cite{wu2014survey,huang2015investigation}.
\citet{caridakis2007multimodal} built a multi-modal discrete emotion classification system based on facial expressions, body movement/gestures, and speech. 
While all fusion methods improved over monomodal classification, feature fusion provided the best performance. 
\citet{hu2017learning} won the EmotiW challenge in 2017 by proposing a novel score function that added model-level fusion at multiple levels within a neural network. 
\cite{mirsamadi2017automatic} obtained state of the art results on the IEMOCAP dataset~\cite{busso2008iemocap} by separately learning spatio-temporal features using a recurrent neural network.
Recently, \citet{schoneveld2021leveraging} combined a recurrent neural network with model-level fusion and reported state-of-the-art results on the RECOLA dataset~\cite{ringeval2013introducing}.
Although integrating audio and video generally increases emotion recognition performance,
a common limitation of all previous methods is that all modalities also need to be present at test time.
However, this is rarely the case in real-world applications.
To address this limitation,~\citet{han2019emobed} proposed EmoBed -- an approach that only required a single modality at test time, yet could still profit from both modalities during training.
To this end, EmoBed aligned video and audio embeddings in a joint space and used a subsequent network to predict emotion labels from embeddings of either modality.
Most recently,~\cite{rajan2021robust} proposed Stronger Enhancing Weaker (SEW), a method that allowed to exploit a stronger modality during training to improve test performance of a weaker modality.
The crucial difference to EmoBed is that this method cannot improve the test performance of a strong modality by training jointly with a complementary but weaker modality.
While these ``asymmetric'' approaches were shown to improve monomodal testing performance, they disregarded gaze for emotion recognition.
In stark contrast, we show that integrating an explicit gaze representation into the visual pipeline of such an asymmetric approach improves test performances both on video- as well as on audio alone.
Given that our literature review did not indicate a clear preference for early- or model level fusion of feature representations, we evaluate the performance of both alternatives of joining gaze features with the visual processing stream.

\section{Method}

\begin{figure}
    \centering
    \includegraphics[width=\textwidth]{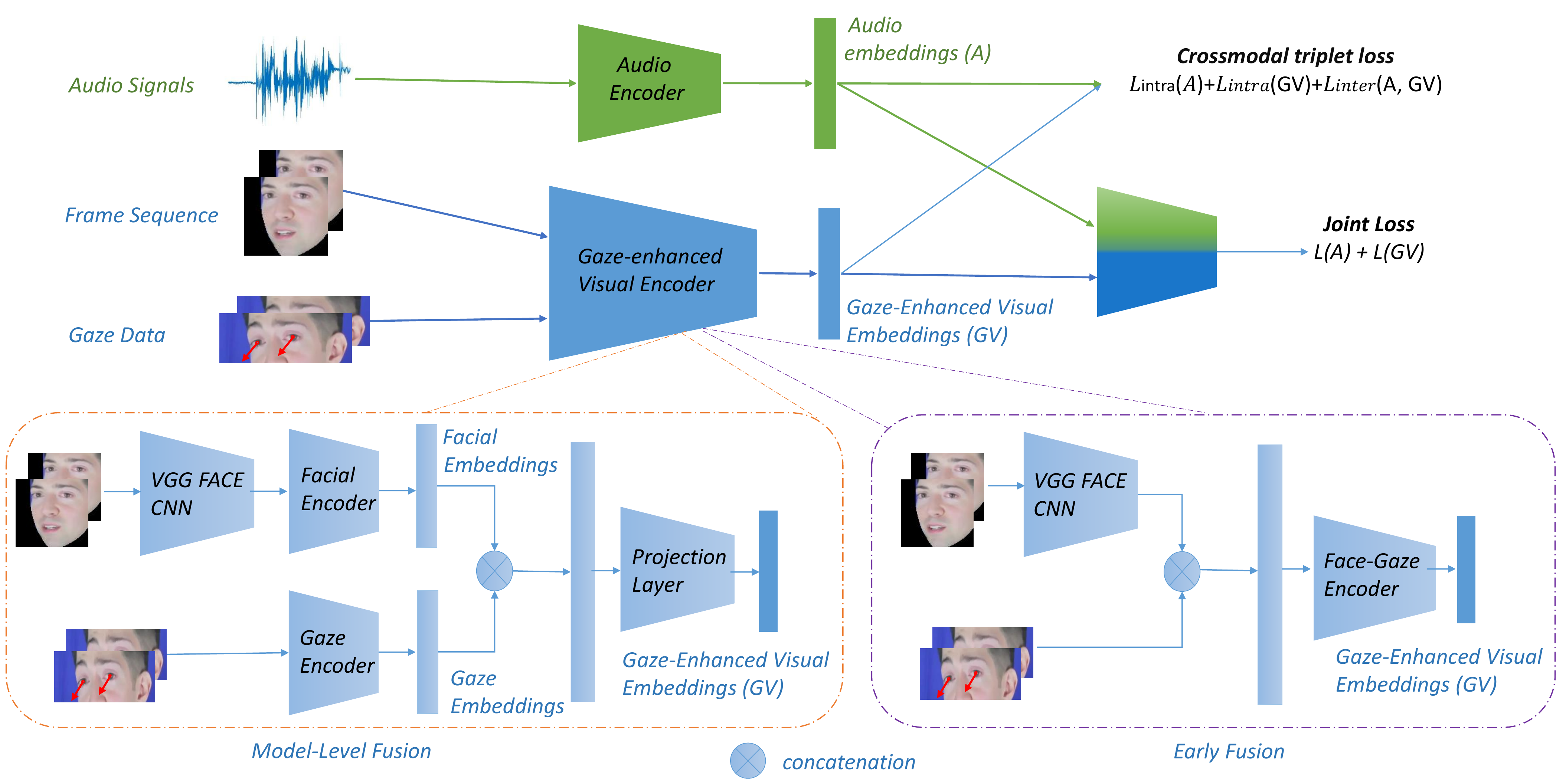}
    \caption{Overview of our proposed gaze-enhanced crossmodal emotion embeddings method.
    \textbf{Top}: general structure of the crossmodal emotion embeddings framework. Audio and video signals are encoded in separate streams. 
    A crossmodal triplet loss is applied to the embedding space to align embeddings from the two modalities.
    For classification, the embeddings are fed to a joint classification network that is agnostic of the embeddings' source modality.
    \textbf{Bottom Left}: Our proposed model-level fusion network for integration of gaze information into the visual encoder. Here, VGG FACE features and gaze features are concatenated after application of the encoder networks.
    \textbf{Bottom right:} Our proposed early-fusion network. Here, gaze features are concatenated and subsequently fed through a joint encoder network.
    }
    \label{fig:architecture}
\end{figure}

Our method extends the crossmodal emotion embedding (EmoBed) \cite{han2019emobed} framework by integrating an explicit gaze representation into the visual pathway (see~\autoref{fig:architecture}).
EmoBed consists of a visual and a speech stream, each of which produce embeddings of the same size.
These pathways are trained in two ways.
First, a downstream network is trained on the embeddings without knowledge of the embeddings' source modalities to predict the ground truth emotion labels.
Second, in a shared embedding space, intra- as well as inter-modal triplet losses are applied to improve and align visual and audio embeddings.
This training setup only requires a single modality at test time.
Due to the inter-modal triplet loss applied to the embedding space, a second modality supplied at training time is able to improve the embeddings of the test modality.

\subsection{Gaze-enhanced Visual Stream}
We propose a novel gaze-enhanced visual stream network for EmoBed (see~\autoref{fig:architecture}).
This network incorporates a dedicated gaze representation into the visual stream of EmoBed, mapping both input features to an embedding space in $\mathbb{R}^{E}$.

\subsubsection{Feature fusion}
\label{sec:feature_fusion}
As from our literature review it remained unclear what is the best way to fuse feature representation for emotion recognition, we study two alternatives: (1) early fusion, and (2) model-level fusion.

In early fusion, we concatenate the visual and gaze data before passing it to the encoder. 
In other words, the early fusion process can be expressed as $e^{early}_{vg} = f_{vg}([x_{v};x_{g}])$ where $f_{vg}(.):\mathbb{R}^{V+G} \rightarrow \mathbb{R}^{E}$. 
$V$, $G$, and $E$ denote the dimension of visual data input, gaze data input, and the embedding size, respectively.

In model-level fusion, we employ a separate encoder for each of the two data streams.
Their corresponding embeddings are subsequently concatenated, forming an embedding of twice the original size (i.e. $2E$). 
Finally, we pass this new embedding into a projection layer to transform it to the original embedding size $E$. 
In other words, the visual encoder can be expressed as  $f_{v}(.):\mathbb{R}^{V} \rightarrow \mathbb{R}^{E}$, the gaze encoder as $f_{g}(.):\mathbb{R}^{G} \rightarrow \mathbb{R}^{E}$, and the projection network as $f_{p}(.):\mathbb{R}^{2E} \rightarrow \mathbb{R}^{E}$.  After passing the visual input $x_{v}$ and gaze input $x_{g}$ into their corresponding encoders we get $e_{v} = f_{v}(x_{v})$ and $e_{g} = f_{g}(x_{g})$. Finally, we project them again into the $\mathbb{R}^{E}$ dimension by $e^{model}_{vg} = f_{p}([e_{v};e_{g}])$.

\subsection{Audio Stream}
Similarly, the goal of the audio stream is to map each audio input to an embedding space of the same dimensionality of the gaze-enhanced visual embeddings (i.e. $\mathbb{R}^{E}$). 
To this end, we feed the audio input $x_{a} \in \mathbb{R}^{M}$ into the audio encoder network which can be expressed as $f_{a}(.):\mathbb{R}^{M} \rightarrow \mathbb{R}^{E}$.
We obtain the resulting audio embeddings $e_{a} = f_{a}(x_{a})$.

\subsection{Triplet Training}
Triplet training is applied on the gaze-enhanced visual embeddings and the audio embeddings.
The goal of the triplet training is to align the two different types of embeddings in a shared embedding space.
The embeddings in the new space, regardless of their input source, should be close to each other if they have the same label, and far from each other otherwise. 
We quantify the semantic similarity using the euclidean distance. 
I.e. the similarity between two embeddings $e_{i}$ and $e_{j}$ is defined as $d(e_{i}, e_{j}) = ||e_{i} - e_{j}||_{2}$.

The triplet loss consists of an intra-modal as well as an inter-modal component. 
The intra-modality loss ensures that embeddings that carry the same label within the same modality are close to each other and far from each other otherwise.
The inter-modality on the other hand ensures that embeddings carrying the same label across modalities are close to each other and far from each other otherwise.

\subsubsection{Intra-modality loss}
 Given a batch of $n$ embeddings $A$ from a single modality, the intra-modality loss is calculated by first computing the $nxn$ pairwise distance matrix. 
 The elements on its diagonal are zero as they represent the distance between each embedding and itself. 
 For each embedding $e\in A$, we followed \cite{han2019emobed} in the triplet mining technique, i.e. we choose the hardest positive- as well as the hardest negative example.
 The hardest positive example $e^{+}$ is defined as the example with the same label that is farthest from $e$ in the embedding space.
 Similarly, the hardest negative example $e^{-}$ is the the example that is closest to $e$ in the embedding space, but has a different label than $e$.
 Then, the intra-modalitiy loss of modality $A$ is defined as as:
\begin{align}
 L_{intra}(A) = \sum^{n} ( d(e_{a},e_{a}^{+}) - d(e_{a},e_{a}^{-}) )   
\end{align}

\subsubsection{Inter-modality loss}
 Given a batch of $n$ examples with both embeddings $e_a$ of modality $A$ and corresponding embeddings $e_b$ of modality $B$, we first compute the cross-modality $nxn$ distance matrix. 
 This matrix contains the distances $d(e_a,e_b)$ for all $e_a\in A$ and $e_b\in B$.
 The elements on its diagonal represent the distances between the $e_{a}$ and $e_{b}$ embeddings belonging to the same video instance. 
 Then, with the same strategy for searching for hardest positives and negatives, the inter-modality loss is calculated by
\begin{align}
 L_{inter}(AB) = \sum^{n} ( d(e_{a},e_{b}^{+}) - d(e_{a},e_{b}^{-}) )   
\end{align}

\subsubsection{Full triplet loss}
For two batches of embeddings $A$ and $B$ of size \emph{n} coming from two different modalities, we calculate the final triplet loss as follows:
\begin{align}
 L_{triplet} = L_{intra}(A) + L_{intra}(B) + L_{inter}(AB)
\end{align}

Thus, regardless of their modality, the triplet loss pulls embeddings that have the same label close to each other while it pushes embeddings with different labels apart.

\subsection{Joint Training}
At the same time when the embedding space is trained with the triplet loss, the embeddings from each modality are passed to a shared classifier. 
This classifier is agnostic to the source modality of the embeddings.
It is trained to classify emotion categories based on the embeddings it receives.
That is, given an embedding $e \in \mathbb{R}^{E}$ the classifier is a function $f(e) = \hat{y}$, where $\hat{y} \in \mathbb{R}$ is the predicted label for $e$.
Thus the total loss of the classifier can be expressed as:
\begin{align}
 L_{joint} = L(A) + L(B)   
\end{align}
where $L(A)$, $L(B)$ are the cross entropy loss for inputs embeddings  $A$ and $B$.

\subsection{Features}

\begin{table}[t]
\centering
\caption{Base features from OpenFace output and the corresponding statistical features. These 103 features are subset of the feature set in \cite{o2019eye}. \textit{LR} refers to a linear regression fitted to the time series of feature values in the window. \textit{Time ratio} is the proportion of time during which a binary feature is detected in the analysis window.
\textit{IQR} denotes the interquartile range, i.e. \textit{IQR 2-3} refers to the difference between third and second quartile.
}
\begin{tabular}{p{4.9cm} p{6.5cm} c}
 \toprule
 Base feature &  Statistical functionals  & \# Features\\
 \midrule
 
 gaze angle x, gaze angle y, $\Delta$ gaze angle x, $\Delta$ gaze angle y, pupil diameter mm
  &
  min, max, mean, median, quartile 1, quartile 3, std, IQR 1-2, IQR 2-3, IQR 1-3, LR intercept, LR slope  & 60\\
 \midrule
$\Delta$ pupil diameter mm &   min, max, mean, quartile 1, quartile 3, std, IQR 1-2, IQR 2-3, IQR 1-3, LR intercept, LR slope & 11\\
\midrule
eye blink intensity & max, mean, median, quartile 3, std, IQR 1-2, IQR 2-3, IQR 1-3, LR intercept, LR slope & 10\\
\midrule
pupil dilation, pupil constriction& time ratio, mean time, max time, total time & 8\\
\midrule
gaze approach & time ratio, mean time, max time, median time & 4\\
\midrule
eyes closed, gaze fixation & time ratio, min time, max time, mean time, median time& 10\\

 \bottomrule
\end{tabular}
\label{table:gaze_features}
\end{table}

\subsubsection{Gaze Features}
As a basis for gaze feature computation, we make use of OpenFace~\cite{baltrusaitis2018openface}. 
OpenFace makes use of a Constrained Local Neural Field (CLNF) landmark detector presented in~\cite{baltrusaitis2013constrained,wood2015rendering} for eyelids, iris, and the pupil detection. 
OpenFace is trained using the SynthesEyes dataset~\cite{wood2015rendering} which contains photorealistic close-up images of eyes and is widely used for training and evaluating gaze estimation models. 
The gaze estimation of OpenFace~\citet{baltrusaitis2018openface} was originally evaluated on the MPIIGAZE dataset~\cite{zhang2015appearance}, which is widely used for appearance based gaze estimation and contains realistic in-the-wild images of users taken from webcams.
OpenFace reached an angular error of 9.1 in the challenging cross-dataset evaluation scenario, clearly outperforming previously proposed methods~\cite{baltrusaitis2018openface}.
Openface was subsequently used for a multitude of tasks, such as mutual gaze prediction for human-robot interaction~\cite{saran2018human}, improving methods for controlling devices with gaze~\cite{kim2019watch}, and anticipating averted gaze~\cite{muller2020anticipating}. 
As such, Openface is a well established and accurate toolkit for facial landmark detection and gaze estimation, particularly for real world person independent gaze estimation. 
While a dedicated eye tracking device can reach higher gaze estimation accuracy, this would limit our approach to highly controlled lab-based scenarios.

Based on the gaze estimates obtained by OpenFace, we compute a set of statistical features from the recently proposed gaze feature set by \citet{o2019eye}. 
We observed that some features of \citet{o2019eye} can have zero variance during a time window, rendering computation of skewness and kurtosis impossible.
As a consequence, we restrict our set of features to those from \citet{o2019eye} that are not subject to this issue.

In particular, we extracted a base set of 9 gaze related features, on top of which we computed several statistical functionals, resulting in 103 features in total. 
The base set consists of 4 numerical and 5 binary features. 
The numerical features are eye gaze direction in radians in world coordinates averaged for right and left eyes, the pupil diameter estimated from the left eye landmarks, and the eye blink intensity (i.e. Facial Action Unit 45\footnote{list of action units and their description https://imotions.com/blog/facial-action-coding-system/}).
The binary features are the pupil dilation time, constriction time, eye closed time, gaze approach time, and gaze fixation time. 
Gaze approach is calculated by taking the derivative of the average depth of the eye landmarks. 
If the derivative is above zero, it is counted as gaze approach.
See \autoref{table:gaze_features} for an overview over all base features and the applied statistical functionals.

OpenFace outputs pupil diameter in millimeters based on the detected facial landmarks~\cite{baltrusaitis2018openface}. 
As we do not assume knowledge of the camera intrinsics, the absolute value of the estimated pupil diameter may be inaccurate, as it is the case in previous work that computed pupil diameter features based on OpenFace output~\cite{o2019eye}.
However, for the task of gaze-based emotion recognition, we are only interested in relative differences in pupil diameter that are preserved even if the absolute value is incorrectly estimated.
We still use ``millimeter'' to be in line with the terminology of~\cite{o2019eye}.

We extract the features on windows of 1 second, resulting in as many feature values per window as there are frames per second in the input video.
These window-based features can be used both in the early- as well as the model-level fusion model discussed in Section~\ref{sec:feature_fusion}, as they can be aligned with the frame-wise facial features.
In the model-level fusion scenario, we also investigate a version of the gaze features where we average the features that were first computed on the window level across the whole utterance.
We suspect this could lead to a more robust gaze feature representation that is less prone to random fluctuations.

\begin{figure}
    \centering
    \includegraphics[width=\textwidth,keepaspectratio]{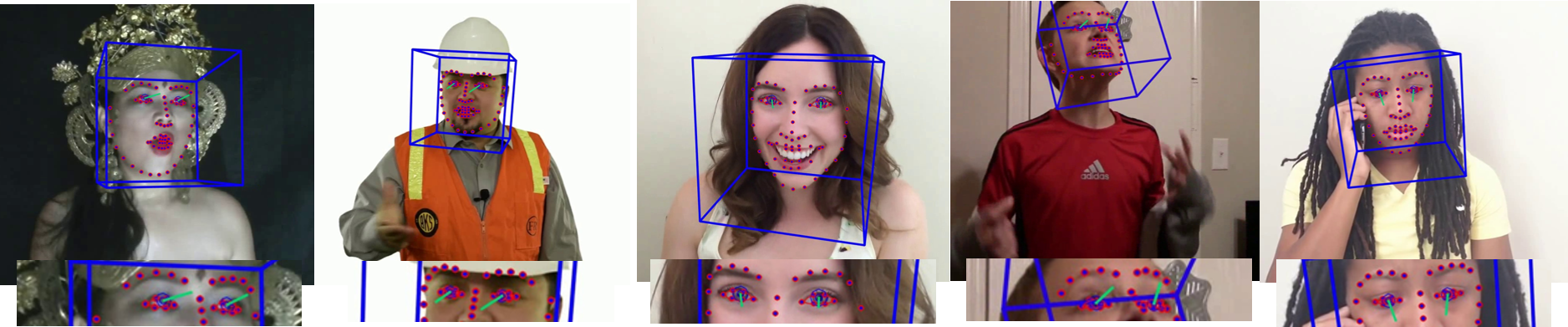}
    \caption{Facial keypoints and gaze estimates for randomly chosen frames with \textbf{high} OpenFace confidence.}
    \label{fig:high_conf_samples}
\end{figure}
\begin{figure}
    \centering
    \includegraphics[width=\textwidth,keepaspectratio]{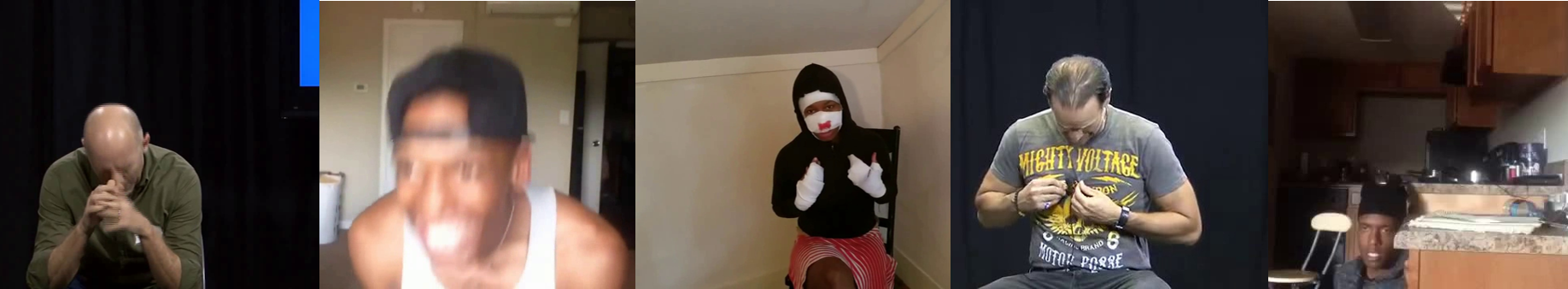}
    \caption{Facial keypoints and gaze estimates fn randomly chosen frames where landmarks detected with \textbf{low} OpenFace confidence}
    \label{fig:low_conf_samples}
\end{figure}

\subsubsection{Facial Features}
We used OpenFace \cite{baltrusaitis2018openface} for face detection and face alignment. 
We pass the aligned face images to the VGG-Face model~\cite{BMVC2015_41} and extract the
the 4096 length feature vector from the "fc-7" layer.
If k is the number of frames in the video, this results in (k,4096) feature values for that video.

\subsubsection{Audio Features}

In line with~\cite{han2019emobed}, we extracted the eGemaps feature set \cite{eyben2015geneva} using the OpenSMILE toolkit \cite{eyben2010opensmile}. This results in a 88-dimensional feature vector per video.

\section{Experimental Evaluation}

\subsection{Training Details}

Inspired by \citet{han2019emobed}, we used GRU-RNN layers for each modality encoder as well as for the shared classification network. 
We used a single hidden layer for both the encoders as well as the shared network with 120 units each. 
We applied L2 regularization of $10^{-4}$ via weight decay, and chose a batch size of 64. 
We found that using the Adam optimizer~\cite{kingma2014adam} with a $10^{-4}$ learning rate yielded the best generalization results on the OMG dataset.
Moreover, all inputs were standardized to their mean and variance on the training set. 
Each batch of input data yielded two batches of the same size:
One batch containing gaze-enhanced visual embeddings and another batch of audio embeddings.
Crossmodal triplet loss is applied on both batches.
Furthermore, the batches were passed in an alternating fashion to the joint classification network. 
That is, the classification network receives the audio embeddings of batch 1, then the gaze-enhanced video embeddings of batch 1, the audio embeddings of batch 2 and so forth.

\begin{table}[t]
\centering
\caption{F1 scores for different approaches and test modalities on the OMG~\cite{barros2018omg} dataset. Scores are averaged across 20 runs with different random intialisations and corresponding standard errors are shown in brackets.}
\begin{tabular}{p{5.3cm} c c c} 
 \toprule
  Approach $\downarrow$ || Test Modality $\rightarrow$ &  Audio & Video\\
 \midrule
 OMG baseline~\cite{barros2018omg} & 33.0 & 37.0 \\
 EmoBed~\cite{han2019emobed} & 41.7 & 43.9 \\ 
 \midrule
 \textit{monomodal}  & & \\
 \ \ \ \ \ no gaze & 41.3 (0.15) & 42.8 (0.24)\\
 \ \ \ \ \ averaging, model-level fusion & - & 43.8 (0.21)\\
 \ \ \ \ \ windowing, early fusion & - & 43.0 (0.23) \\
 \ \ \ \ \ windowing, model-level fusion & - & 43.7 (0.20) \\
 \midrule
 \textit{crossmodal triplet training} & & \\
 \ \ \ \ \ no gaze & 42.9 (0.17) & 43.5 (0.25) \\
 \ \ \ \ \ averaging, model-level fusion & 42.6 (0.19) & \textbf{45.0} (0.22)\\
 \ \ \ \ \ windowing, early fusion & \textbf{43.4} (0.17) & 43.7 (0.26)\\
 \ \ \ \ \ windowing, model-level fusion & 42.8 (0.15) & 44.5 (0.20) \\
\bottomrule
\end{tabular}

\label{table:main_results}
\end{table}

\subsection{Dataset}

We evaluated our gaze-enhanced crossmodal triplet training on the popular One-Minute Gradual Emotional (OMG) Behavior dataset \cite{barros2018omg}. 
We opted for this dataset given that it is well suited for the task and has been used in closely related prior work \cite{han2019emobed}.
The OMG dataset~\cite{barros2018omg} consists of 567 monologue videos collected from YouTube with an average video length of one minute and a frame rate of 29 fps.  
Each monologue video contains various emotional behaviors shown by the person in the video.
Each video was subdivided into several utterances with an average length of 8 seconds. 
The utterances were subsequently annotated by 5 different Amazon Mechanical Turk (AMT) workers each with discrete labels for the six basic emotions according to \citet{ekman1971constants}. 
Addition of a neutral class resulted in the seven classes: \textit{disgust}, \textit{anger}, \textit{fear}, \textit{happy}, \textit{neutral}, \textit{sad}, and \textit{surprise}.
Annotators saw the utterances in sequence, i.e. they could take into account context information from preceding utterances of the same video.
As annotations are based on the full array of human behavior channels including speech, gaze, and facial expressions, this dataset is well suited to evaluate our gaze-enhanced crossmodal embedding approach.

Before feature extraction, we removed four utterances from the training set for which neither OpenFace could detect any face nor OpenSMILE could detect any audio signal. 
This resulted in a training dataset of 2,438 utterances in training and 617 utterances in the validation set.
All experiments and results reported in the following have been performed on the validation set given that the test labels are not available for OMG.
Using the OMG validation set allows us to directly compare our results with those of 
\cite{han2019emobed}.

The facial landmark detector of OpenFace 2.0~\cite{baltrusaitis2018openface} reported a confidence score of more than 0.9 for 91.2\% of all frames in the OMG dataset, indicating that in the vast majority of frames high-quality gaze estimates can be expected.
For a qualitative analysis we randomly sampled one frame for each gaze direction (left, right, up, down, straight) from the set of frames with confidence score of at least 0.9 (see Figure~\ref{fig:high_conf_samples}).
All sampled frames show accurate estimates of gaze direction.
Figure \ref{fig:low_conf_samples} shows five frames where facial landmark detection failed. 
Such complete failures occur seldom (2.5\% on the training set) and are often due to occlusions, motion, or extreme head poses.

\begin{table}[t]
\centering 
\caption{Average of F1 scores for each emotion class for cross modal triplet training. Standard errors across 20 runs are shown in brackets.}
\label{table:performance_classes}
\begin{tabular}{p{3.5cm} c c c c c c c c} 
 \toprule
   Approach $\downarrow$ || Emotion $\rightarrow$ 
   &Neutral & Happy & Sad & Anger & Disgust & Fear & Surpr. \\
   \midrule
   Train Samples& 888 & 713 & 346 & 293 & 112 & 60 & 26\\
   Validation Samples & 196 & 207 & 79 & 61 & 48 & 18 & 8 \\
  \midrule
  \textit{test modality: video}\\
  \ \ \ \ \ averaging, &
  49.6(0.5) & 55.7(0.4) & 32.7(0.8) & 29.2(1.3) & 4.7(1.3) & 0 & 0\\
  \ \ \ \ \ \ \ \ model-level fusion \\
  \ \ \ \ \ no gaze& 48.1(0.6) & 52.3(0.4) &29.7(1.5) & 32.5(1.5)& 11.6(1.7) & 0  & 0 \\
  \textit{test modality: audio}\\
  \ \ \ \ \ windowing, & 49.7(0.4) & 49.7(0.5)  & 34.8(0.8) & 25.3(1.5)& 0 & 0&0\\
  \ \ \ \ \ \ \ \ early fusion \\
  \ \ \ \ \ no gaze& 49.2(0.4)& 49.8(0.5) & 33.9(0.8)& 23.4(1.1)& 0 & 0   & 0 \\

\bottomrule
\end{tabular}
\end{table}

\subsection{Results}
 
In line with previous work~\cite{barros2018omg,han2019emobed}, we use the micro F1 score as our general performance metric.
It is calculated by counting the total true positives, false negatives and false positives globally across all classes.
\subsubsection{Effect of Gaze Integration}
We compare our proposed method with the previous state of the art reported by \citet{han2019emobed} as well as ablated versions in \autoref{table:main_results}.
For video as test modality, we achieve 45.0 F1 for our crossmodal triplet training method that uses averaged gaze features and model-level fusion.
This is a clear improvement over the previous state of the art of 43.9 F1.
In the case of audio-only testing, our crossmodal triplet training method with gaze features extracted on windows and early fusion achieves the best result with 43.4 F1.
Again, this outperforms the previous state of the art of~\citet{han2019emobed} at 41.7 F1.

At the same time, our best gaze integration methods for each test modality improve over their corresponding no gaze ablation at 43.5 F1 for video testing, and 42.9 F1 for audio testing.
The ``no gaze'' ablation represents our implementation of the EmoBed framework.
For audio-only testing, it improves over the performance originally reported in \citet{han2019emobed} by 1.2 F1, and for video only testing it is worse by 0.2 F1.
While all gaze integration strategies improve over the no gaze triplet training alternative for video testing, the results are more mixed for audio-only testing.
Here, only windowed gaze extraction combined with early fusion improves over the no gaze ablation.

In addition to crossmodal training setups, we investigate purely monomodal training.
Most importantly, all our crossmodal triplet training models improve over their monomodal counterparts.
Furthermore, the pattern of results for different gaze integration strategies in monomodal video models follows the corresponding pattern of results for our crossmodal triplet training models.
This indicates that the advantage of averaging and model-level fusion for video-only testing is stable across evaluation conditions.

\subsubsection{Analysis of improvements per emotion class}
To gain a deeper insight into the performance of our gaze integration method, we conducted a performance analysis per emotion class. 
In~\autoref{table:performance_classes}, we compare the performances of our gaze-enhanced crossmodal triplet training approaches to the corresponding no-gaze ablations. 
For video testing, we observe an improvement for our the gaze-enhanced method for the classes neutral (+1.5 F1), happy (+3.4 F1), and sad (+3.0 F1).
On the other hand, our gaze-enhanced method obtains a lower F1 score for anger (-3.3 F1) and disgust (-6.9 F1).
The improvements are aligned with the number of training samples available for each class.
Our gaze-enhanced approach improves for classes with high numbers of training samples and has a disadvantage on classes with low numbers of training samples.
A different pattern can be observed in the audio testing case.
Here, our gaze-enhanced approach improves for the classes neutral (+0.5 F1), sad (+0.9 F1), as well as anger (+1.9 F1), and is on par with the no gaze baseline for happy (-0.1 F1).
These varied behaviors point to the different roles gaze features play when incorporated into the helper modality (audio testing), or when used as a direct basis of test time prediction in video testing (see our Discussion in Section~\ref{sec:discussion_performance}).

Interestingly, all approaches only reach a very low performance on classes with a low number of training samples (i.e. disgust, fear, and surprise).
To check whether this might be due to not yet converged training for these emotion classes, we tried training our model for a large number epochs (250).
This did not resolve the issue but lead to overfitting.

\subsubsection{Comparing Gaze Feature sets}
\begin{table}[t]
\centering
\caption{F1 scores of different gaze feature representations in the crossmodal triplet training formulation on the OMG~\cite{barros2018omg} dataset.
Standard errors across 20 runs are shown in brackets.}
\begin{tabular}{p{5.3cm} c c c c} 
 \toprule
   Gaze Features $\downarrow$ || Test Modality $\rightarrow$ & Audio & Video \\
  \midrule
  \textit{\citet{o2019eye}} \\
    \ \ \ \ \ averaging, model-level fusion & 42.6 (0.19) & \textbf{45.0} (0.22) \\
  \ \ \ \ \ windowing, early fusion & \textbf{43.4} (0.17) & 43.7 (0.26) \\
  \ \ \ \ \ windowing, model-level fusion & 42.8 (0.15) & 44.5 (0.20) \\
  \midrule
  \textit{\citet{van2019emotion}} \\
    \ \ \ \ \ averaging, model-level fusion & 43.1 (0.16) & 44.8 (0.24)\\
  \ \ \ \ \ windowing, early fusion & 43.2 (0.18) & 43.6 (0.28) \\
  \ \ \ \ \ windowing, model-level fusion & 42.8 (0.18) & 43.5 (0.23) \\

\bottomrule
\end{tabular}
\label{table:gaze_comparison}
\end{table}

Our gaze-enhanced visual stream utilizes a set of gaze features proposed for emotion recognition by \citet{o2019eye}.
A different set of features was proposed by \citet{van2019emotion}, which to the best of our knowledge, was the only other feature set proposed in the emotion recognition literature that utilizes gaze estimates extracted from standard RGB videos (e.g. via OpenFace~\cite{baltrusaitis2018openface}).
The main difference between these feature sets is that the features of \citet{van2019emotion} do not contain information about the pupil dilation and constriction, nor eye blinking.
On the other hand, they include information about the iris and the average location of the eye landmarks.
In detail, 51 statistical features were calculated from a base set of features extracted by OpenFace.
These base features include the eye gaze direction vector in 3D and radians averaged for both eyes, height and width of the pupil for both eyes, the eye locations, as well as the vertical distances of the iris landmarks.

\autoref{table:gaze_comparison} shows the results of a comparison of our gaze featureset inspired by \citet{o2019eye} with the one by \citet{van2019emotion} in the crossmodal triplet training approach.
In both audio and video testing scenarios, the best performance is achieved using the featureset based on \citet{o2019eye}.
While for video-only testing, the features of \citet{o2019eye} are always superior to those of \citet{van2019emotion}, the results for audio-only testing are more mixed.

\subsubsection{Gaze Feature Analysis}

\begin{figure}
    \centering
    \includegraphics[width=\textwidth,keepaspectratio]{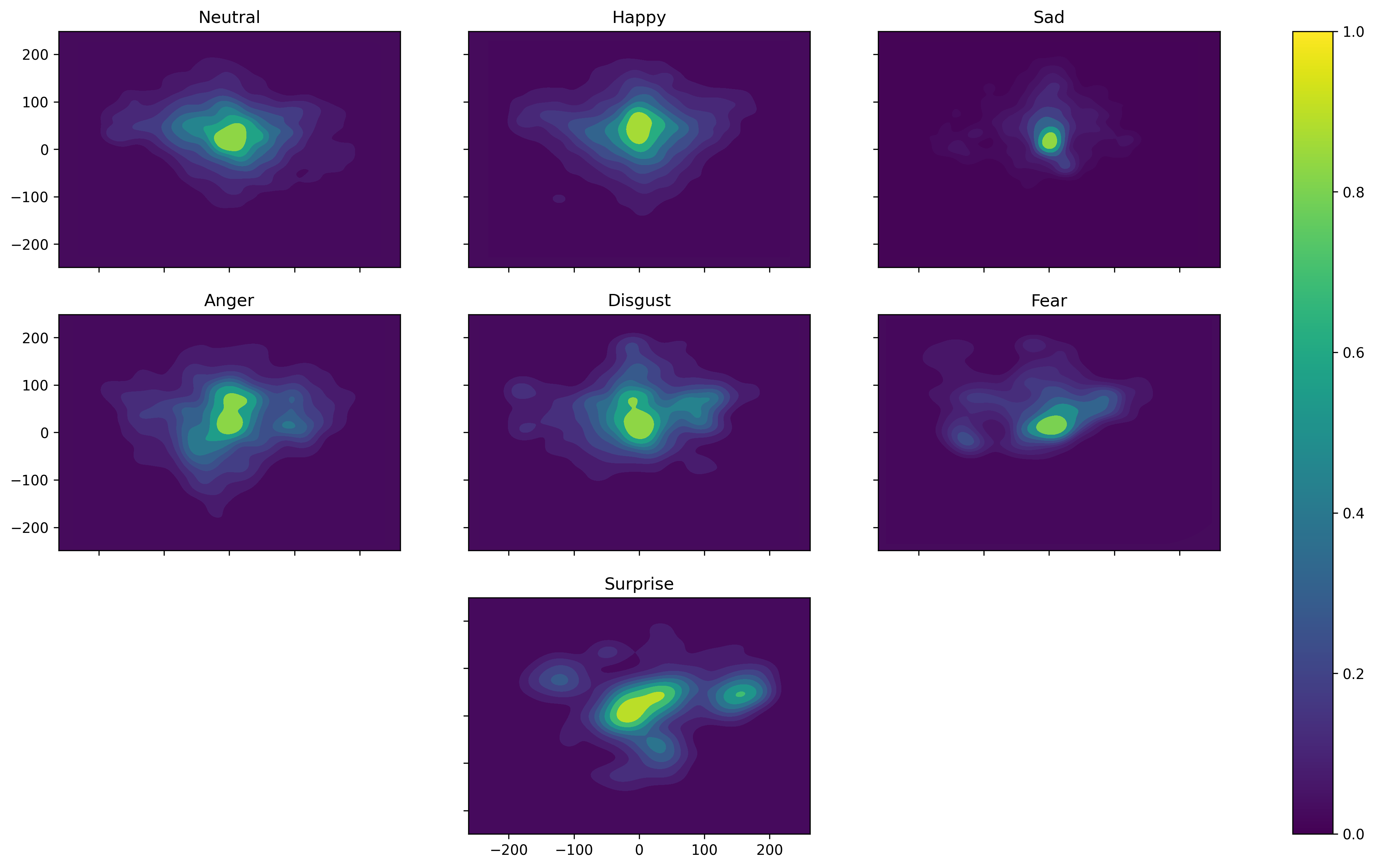}
    \caption{Visualization of the distribution of OpenFace gaze estimates projected on the camera plane for different emotion classes.}
    \label{fig:kde_plot}
\end{figure}

We visualized the distribution of gaze estimates for each emotion class by projecting them on the camera plane.
\autoref{fig:kde_plot} shows the kernel density estimation of these projections, demonstrating the connection between gaze estimates and the emotions expressed in the OMG dataset. For instance, the gaze distribution for Sadness is very concentrated which is in line with passivity that is associated with this emotion. Surprise on the other hand is more spread out.

We additionally performed a feature importance analysis for each emotion class. 
We used maximum relevance minimum redundancy (mRMR)~ \cite{ding2005minimum} to rank the features that would most distinguish this class from the others in a one versus all manner. 
The top 5 features for each class are shown in Table \ref{tab:mRMR_label}. 
The dominance of the gaze angle based features is in line with previous results on feature importance analysis for arousal and valence regression~\citet{o2019eye}. 
The gaze angle based features, which represent 46.6\% of the full feature set, contributed to 71.4\% of the top five ranked features in Table~\ref{tab:mRMR_label}. 
Furthermore, pupil diameter, eye blink intensity, and gaze approach account for 10.6\%, 9.7\%, and 3.8\% of the full feature set, while contributing to the top ranked features by 17.1\%, 8.5\%, and 2.8\% respectively.

\section{Discussion}

\subsection{Achieved Performance}
\label{sec:discussion_performance}

In our experiments, gaze integration into the visual stream of the crossmodal emotion embeddings (EmoBed)~\cite{han2019emobed} framework clearly improved test performances for both video-only as well as audio-only testing.
The precise gaze integration method affected performances differently for different test modalities, i.e. we did not find a single best gaze integration method that is superior across all scenarios.
When testing on video, we reached the best performance when averaging gaze features across the whole input videos and performing model-level fusion.
In contrast, when testing on audio, we achieved best performance for window-based features and early fusion.
This difference might be the result of the different roles gaze features play in each of these test scenarios.
In the visual testing scenario, prediction is performed directly on gaze inputs.
A smaller feature representation that is less subject to noise (i.e. averaged across the whole input video) might help in this scenario.
On the other hand, when gaze features are part of the helper modality, a certain degree of randomness might be beneficial as it can act as a regularizer in the embedding space.
This interpretation is in line with our analysis of performances achieved in each emotion class (\autoref{table:performance_classes}).
Here, in contrast to the overall improvement, gaze integration for video testing decreased performance for emotion classes with a low number of samples (disgust and anger).
The combination of the low number of samples with the higher number of features introduced due to gaze integration might be the reason for impaired performance.
In the audio testing scenario however, we did not observe a decrease in performance for classes with a low number of samples, supporting our reasoning that a larger feature space of the helper modality might have less detrimental effects.

Our class-specific performance analysis in \autoref{table:performance_classes} also revealed that both our proposed gaze-enhanced approach as well as our no gaze baseline implementation of EmoBed~\cite{han2019emobed} achieved an F1 score of 0 on fear and surprise.
These low performances are clearly related to the low number of training examples available for these classes.
The consensus performance measurement on the OMG dataset is micro F1~\cite{barros2018omg,han2019emobed}.
We choose this measurement in our evaluations for comparability with previous work.
To improve performance on fear and surprise, a different evaluation metric (e.g. macro F1), and correspondingly, a different loss formulation giving more weight to those rare classes would be needed.

\begin{table}[t]
\centering
\caption{
Top five features selected by mRMR for each emotion class. 
}
\begin{tabular}{p{1.5cm} p{1.5cm} p{1.5cm} p{1.5cm} p{1.5cm} p{1.5cm} p{1.5cm}}
\toprule
Neutral & Happy  &  Sad  & Anger &   Fear &  Disgust & Surprise  \\
\midrule
\small{gaze angle y median}&
\small{gaze angle y quartile 3} &
\small{gaze angle x SD} &
\small{$\Delta$ gaze angle x max}&
\small{gaze angle y LR intercept}&
\small{$\Delta$ gaze angle y quartile 1}&
\small{gaze angle x IQR 2 3}
\\ 
\small{$\Delta$ pupil diameter mm LR slope} &
\small{gaze angle x median}&
\small{$\Delta$ pupil diameter m mean}&
\small{$\Delta$ pupil diameter mm mean}&
\small{$\Delta$ gaze angle x LR slope}&
\small{$\Delta$ pupil diameter mm LR intercept}&
\small{$\Delta$ gaze angle x LR slope}
\\ 
\small{$\Delta$ pupil diameter mm IQR 2-3}&
\small{$\Delta$ gaze angle y median }&
\small{$\Delta$ gaze angle y LR intercept}&
\small{$\Delta$ gaze angle x LR intercept}&
\small{$\Delta$ gaze angle y quartile 3}&
\small{eye blink intensity LR slope}&
\small{$\Delta$ pupil diameter mm LR intercept}
\\ 
\small{eye blink intensity median} &
\small{$\Delta$ gaze angle x IQR 2-3}&
\small{$\Delta$ gaze angle y quartile 3}&
\small{$\Delta$ gaze angle y LR intercept}&
\small{gaze angle x LR intercept}&
\small{gaze angle y quartile 1}&
\small{$\Delta$ gaze angle x median}
\\ 
\small{$\Delta$ gaze angle x LR intercept} &
\small{gaze approach time secs mean} &
\small{gaze angle x IQR 1-3} &
\small{$\Delta$ gaze angle x IQR 1-3} &
\small{eye blink intensity median}&
\small{gaze angle y IQR 1-2}&
\small{$\Delta$ gaze angle y min}
\\
\bottomrule
\end{tabular}

\label{tab:mRMR_label}
\end{table}

\subsection{Limitations and Future Work}
While our work showed the importance of gaze integration in flexible emotion recognition systems, it is still subject to several limitations that should be addressed in future work.

A key prerequisite for every gaze integration method are the available gaze estimates.
While our system was able to clearly improve over no-gaze baselines already with the imperfect estimates provided by OpenFace~\cite{baltrusaitis2018openface}, it potentially missed subtle gaze cues that are not well represented in these gaze estimates.
More accurate gaze estimates obtained from dedicated eye tracking systems~\cite{kassner2014pupil,dowiasch2020quantitative} should be examined in future work as they have the potential to improve performance even further.
Another limitation lies in the dataset we used for training and evaluation.
The OMG dataset~\cite{barros2018omg} consists of monologue videos collected from YouTube.
While this covers a wide variety of different participants and personalities, it poses constraints on the recording situation (i.e. a single person in front of the camera).
Future work should investigate how our approach can be extended to handle less constrained scenarios.
One example is the recognition of emotions embedded in interactions of freely moving people~\cite{muller2015emotion}.
This poses several additional challenges.
For example, gaze estimates might not always be available due to occlusions. 

While we used recent state-of-the-art gaze features for emotion recognition~\cite{o2019eye,van2019emotion}, these featuresets do not represent the target of a persons' gaze.
An important avenue for future work on gaze-based emotion recognition will be to investigate ways to integrate information on the gaze target.
This is especially relevant in scenarios with multiple possible gaze targets, e.g. in group interactions~\cite{muller2021multimediate}.
Such an approach would require eye contact detection~\cite{fu2021using,siegfried2021visual} during pre-processing.

Finally, the principle of gaze integration presented 
is also applicable to tasks beyond emotion recognition in which gaze plays a prominent role.
Examples for such tasks include emergent leadership detection~\cite{muller2019emergent,capozzi2019tracking}, prediction of speaking turns~\cite{birmingham2021group}, rapport estimation~\cite{mueller18_iui}, or personality prediction~\cite{hoppe2018eye}.

\subsection{Applications}
Our method can be applied in any HCI scenario where an RGB video of adequate quality to perform gaze estimation is available.
OpenFace 2.0~\cite{baltrusaitis2018openface} was successfully evaluated in comparison with state-of-the-art gaze estimation approaches and subsequent work has made use of it in a variety of settings, including group interactions recorded from ambient cameras~\cite{fu2021using} and interviews conducted via teleconferencing~\cite{muller2020anticipating}.
As a result, we expect our method to be applicable to a wide variety of situations in which emotion recognition is desired.
Emotion recognition is key in many applications, including telemedicine~\cite{kallipolitis2021speech}, educational technology~\cite{bahreini2016towards}, and artificial mediators~\cite{muller2021multimediate}.
For example, the emotional responsiveness of a psychotherapy patient to her therapist can be used to estimate therapeutic alliance and potentially predict therapy outcome~\cite{koole2016synchrony}.
Gaze-enhanced cross-modal emotion embeddings are promising in this application, especially when unstable network connections result in low-quality or dropped video.
In addition a large number of health-related interactions take place on the phone, without any video availability~\cite{konig2021measuring,troger2018telephone}.
This is an ideal application scenario where speech-based emotion recognition at test time can profit from gaze-enhanced cross-modal emotion embedding training.
In an educational setting, emotion recognition can 
be valuable to access students' evaluation or learning material~\cite{chen2011using}.
As students are often silent during learning or may be in a location with high ambient noise, gaze-enhanced cross-modal emotion embeddings can potentially be used to create video-only emotion recognition systems that can still profit from speech data during training.

\section{Conclusion}
In this work, we presented the first approach for gaze integration into a crossmodal training framework that allows single-modality emotion recognition at test time.
We introduced a new module visual stream network to the crossmodal emotion embedding (EmoBed) framework.
This visual stream network combines gaze features with a facial feature representation.
With a model-level fusion approach of gaze and facial features, we outperformed the state of the art for video only testing on the popular One-Minute Gradual Emotional (OMG) Behavior dataset.
Using an early fusion approach we also improved over the previous state of the art for audio-only testing on the same dataset.
In ablation experiments, we showed that crossmodal training with the gaze-enhanced visual stream network leads to clear improvements over baselines without gaze integration.
Taken together, our results underline the importance of gaze integration in practical emotion recognition systems.

\begin{acks}
A. Bulling were funded by the European Research Council (ERC; grant agreement 801708). E. Sood was funded by the Deutsche Forschungsgemeinschaft (DFG, German Research Foundation) under Germany's Excellence Strategy - EXC 2075 - 390740016. P. M\"uller was funded by the \grantsponsor{01IS20075}{German Ministry for Education and Research (BMBF)}{https://www.bmbf.de/}, grant number \grantnum{01IS20075}{01IS20075}. We would like to thank Dominike Thomas for her support and encouragement during the course of the project. We Thank The German University in Cairo (GUC) for the initial support to A. Abdou.
\end{acks}
\bibliographystyle{ACM-Reference-Format}
\bibliography{References}


\end{document}